\documentclass[conference]{IEEEtran}
\IEEEoverridecommandlockouts
\usepackage{cite}
\usepackage{amsmath,amssymb,amsfonts}
\usepackage{algorithmic}
\usepackage{graphicx}
\usepackage{textcomp}
\usepackage{amssymb}
\usepackage{amsmath}
\usepackage{flushend}
\usepackage{hyperref}

\DeclareMathOperator*{\argmin}{arg\,min}

\usepackage{tikz}
\usepackage{lipsum}

\newcommand\copyrighttext{%
  \footnotesize 978-1-5386-5541-2/18/\$31.00~\copyright2018 IEEE. Personal use of this material is permitted.
Permission from IEEE must be obtained for all other uses, in any current or future 
media, including reprinting/republishing this material for advertising or promotional 
purposes, creating new collective works, for resale or redistribution to servers or 
lists, or reuse of any copyrighted component of this work in other works. 
DOI: \href{https://ieeexplore.ieee.org/document/8498134}{10.1109/ICAC.2018.00023}}
\newcommand\copyrightnotice{%
\begin{tikzpicture}[remember picture,overlay]
\node[anchor=south,yshift=10pt] at (current page.south) {\fbox{\parbox{\dimexpr\textwidth-\fboxsep-\fboxrule\relax}{\copyrighttext}}};
\end{tikzpicture}%
}

\def\paraspace{\vspace{1em}}
\def\BibTeX{{\rm B\kern-.05em{\sc i\kern-.025em b}\kern-.08em
    T\kern-.1667em\lower.7ex\hbox{E}\kern-.125emX}}
\begin{document}

\title{Preparing for the Unexpected:\\Diversity Improves Planning Resilience\\in Evolutionary Algorithms}

\author{\IEEEauthorblockN{Thomas Gabor}
\IEEEauthorblockA{LMU Munich\\thomas.gabor@ifi.lmu.de}
\and
\IEEEauthorblockN{Lenz Belzner}
\IEEEauthorblockA{MaibornWolff\\lenz.belzner@maibornwolff.de}
\and
\IEEEauthorblockN{Thomy Phan}
\IEEEauthorblockA{LMU Munich\\thomy.phan@ifi.lmu.de}
\and
\IEEEauthorblockN{Kyrill Schmid}
\IEEEauthorblockA{LMU Munich\\kyrill.schmid@ifi.lmu.de}
}

\maketitle

\copyrightnotice

\begin{abstract}
As automatic optimization techniques find their way into industrial applications, the behavior of many complex systems is determined by some form of planner picking the right actions to optimize a given objective function. In many cases, the mapping of plans to objective reward may change due to unforeseen events or circumstances in the real world. In those cases, the planner usually needs some additional effort to adjust to the changed situation and reach its previous level of performance. Whenever we still need to continue polling the planner even during re-planning, it oftentimes exhibits severely lacking performance. In order to improve the planner's resilience to unforeseen change, we argue that maintaining a certain level of diversity amongst the considered plans at all times should be added to the planner's objective. Effectively, we encourage the planner to keep alternative plans to its currently best solution. As an example case, we implement a diversity-aware genetic algorithm using two different metrics for diversity (differing in their generality) and show that the blow in performance due to unexpected change can be severely lessened in the average case. We also analyze the parameter settings necessary for these techniques in order to gain an intuition how they can be incorporated into larger frameworks or process models for software and systems engineering.
\end{abstract}


\section{Introduction}

As automatic optimization in various forms makes its way into industrial systems, there is a wide range of expectations about the upcoming capabilities of future ``smart systems'' \cite{harman2012search,de2013software,wirsing2015software,DBLP:journals/corr/AmodeiOSCSM16,thoben2017industrie}. For most of the current applications, the optimization part of the system takes place \emph{offline}, i.e., not while the application is actually performing its main purpose: The product shipped to the customer is fixed after initial training and does not self-adapt (anymore). Instead, it may only gather data that is then used at the vendor's side to either improve the product's performance via software updates later on or assist in building the product's successor. This, of course, misses out on interesting applications that may highly benefit from further optimization even while they are running. In this paper, we focus on the exemplary case of a layout configuration for the positioning of work stations inside a (smart) factory: Depending on the products that need to be build and depending on the current status of the machines involved, we may desire different workflows for the same product at different times during the factory's life. For most current factories, however, the arrangement of workstations is planned far in advance and then fixed until human intervention.

One of the reasons for opting for offline adaptation is that the vendor usually has access to more computational power and that the employed adaptation process can benefit from connecting data input from a variety of customers. However, increasing computational resources and online connectivity mitigate these issues. A possibly more important aspect is the issue of consistent performance: An online planner, while theoretically able to react to sudden changes in its environment and/or objective, may take some time to reach good plans and during that time the solutions provided by the planner may be unsuitable. 

\paraspace{}
\paragraph{Expected Change} The usefulness and importance of \emph{self-optimization} at the customer's side has already been claimed in the original vision of autonomic computing \cite{kephart2003vision} and has been shown on many occasions since \cite{wirsing2015software,chen2015requirements,arcaini2015modeling}. In these cases, self-optimization usually refers to a process of specialization, i.e., the system is built with a large variety of possible use cases in mind and learns to work best for the few of these it actually faces on site. Intuitively, we may want to build a planner that works on factory layouts in general and that can then specialize on the specific needs of a single factory or a single situation (machine failure, e.g.) if necessary. We expect this approach to work iff every possible situation and every pair of follow-up situation is considered when evaluating a factory layout. As long as we know that machines might fail with a certain probability, we can take this into account and plan redundantly with respect to machine usage. This is what we call \emph{expected change} of the evaluation function.

\paraspace{}
\paragraph{Unexpected Change} Still, we may not want our self-optimizing planner to completely break on any deviation from the specified scenarios. We imagine that intelligent planners should invest a certain amount of effort to think about and prepare for ``what ifs'', even when the respective scenarios have not been expected to happen during system design or training. This is further motivated by the fact that many industry applications require the adaptive component to produce a solution better than a certain quality threshold but do not benefit as much from the system finding configurations that are just slightly better beyond that threshold. Instead, that computational effort might be better put into finding alternative solutions that might not be just as good as the primary solution that was just found, but then again might be feasible even when the primary solution fails for some \emph{unexpected} reason.

This argument falls in line with the claim of \emph{self-protection} for autonomic systems \cite{kephart2003vision}: Our system should not only be able to react and recover from negative external influences but also spend a reasonable effort on actively preparing for negative events. Via this self-protection property we aim to increase the overall resilience of the planning process and by extent the robustness of the system using our planner.

\paraspace{}
\paragraph{Scope of This Work} As the original contribution of this paper we identify that diversity in evolutionary algorithms, which we consider a primary example for a heuristic optimization algorithms in this paper, is of central importance for the algorithm's reaction to change and that explicitly optimizing for diversity helps to prepare for changes, even when they cannot be foreseen by the optimization process in any way. We introduce means to formally define the phenomenon of unexpected change in relation to an online planner.

To this end, we first formally define the notions of change and unexpectedness that we used intuitively until now (Section~\ref{sec:found}). We then immediately turn to an example of a smart factory domain in which unexpected change might occur and specify our experimental setup (Section~\ref{sec:experiment}). We introduce our approach at maintaining diversity using two different diversity metrics (Section~\ref{sec:approach}) and sum up the results of applying this approach in the previously defined experiment (Section~\ref{sec:results}) before we discuss related work (Section~\ref{sec:rel}) and conclude this paper (Section~\ref{sec:conclusion}).

\section{Foundations}
\label{sec:found}

We assume that to realize modern challenges in industry, software products need to feature a certain degree of \emph{autonomy}, i.e., they feature at least one component called \emph{planner} capable of making decisions by providing a plan of actions which the system is supposed to perform to best fulfill its intended goal \cite{arcaini2015modeling,belzner2016software}. This goal is encoded by providing the system with a \emph{fitness function} that can be used to evaluate plans. A planner respecting a fitness function performs self-optimization.

We claim that for many real-world applications it is often not only important to eventually adapt to new circumstances but also to avoid causing any major damage to overall success while adapting. It follows that the planner needs to offer a suitable solution at all times, even directly after change in the environment. This property can be compared to the \emph{robustness} of classical systems, i.e., the ability to withstand external changes without being steered away too far from good behavior \cite{bankes2010robustness}. Robustness can often be tested against a variety of well-defined external influences. However, not every influence a system will be exposed to can be foreseen.\footnote{When possible, endowing systems with means to perceive all possible influences and events might be highly beneficial to resilience. We work with the assumption that this is not always possible or feasible.} The notion of \emph{resilience} captures the system's ability to withstand \emph{unanticipated} changes \cite{florio2013constituent}.\footnote{It follows that we consider resilience a special instance of robustness: Robustness may include both anticipated and unanticipated change. Resilience focuses on the latter.} One approach to prepare a system for unexpected circumstances is to make it adapt faster, so that its adaptive component finds a new plan of actions faster once the old one is invalidated. However, this approach is still purely reactive and we thus cannot prevent the immediate impact of change.

To increase system resilience, we thus might want the planner to become proactive towards possible changes that may occur to the environment and by extension the planner's objective. In order to lessen the blow of unexpected changes, the planner thus needs to prepare for it before it actually occurs. Note that for the changes we are talking about in this section, we still assume that they are unexpected at design time. The planner therefore has no means of predicting when or what is going to happen. Still, we desire for a planner to be caught off-guard as seldom as possible. A planner that needs to re-plan less often would then be considered more resilient with respect to unexpected change. We claim that explicitly increasing planning resilience aids a system's ability to self-protect and is thus a useful handle to explicitly expose to the developers of such a system.

\paraspace{}
\paragraph{Planning} Planners perform (usually stochastic) optimization on the system's behavior by finding plans that (when executed) yield increasingly better results with respect to a specified objective. That objective is given via a fitness function $f: P \times E \to \mathbb{R}$, where $P$ is the domain of all possible plans and $E$ is the domain of environments said plans are to be executed in. For the purpose of this paper, we assume that we want to \emph{minimize} the real-valued output of the fitness function. We can then describe a planner formally as a function $\textit{plan}: E \to P$ from an environment $e \in E$ to a plan $p \in P$ with the following semantic:
$$\textit{plan}(e) \approx \argmin_{p \in P} \mathbb{E}(f(p, e)).$$
Note that due to the possibly stochastic nature of the environment and in extent the evaluation of the fitness function $f$, we compute the expected value $\mathbb{E}$ of the application of $f$. Further note that due to the stochastic nature of the planning methods considered in this paper, we may not actually return the single best result over the domain of all plans but when the stochastic optimization process works, we expect to yield a result somewhat close (described by $\approx$). To compute a reasonable value for $f(p, e)$, a given plan will usually be executed in a simulated version of $e$. We call the process of repeatedly calling $\textit{plan}$ to execute the currently best solution \emph{online} planning, which implies that we may call it for changing~$e$.



\paraspace{}
\paragraph{Changing Environments} We can write any occurrence of change in the environment as a function $c : E \to E$. Obviously, if we allow any arbitrary change to happen to the environment, we can construct arbitrarily ``evil'' environments and cause the planner to perform arbitrarily bad. But frankly, we do not care for a planner managing a smart grid's power production to perform well when a meteor destroys Earth. What is much more realistic and thus much more desirable to prepare for, however, is changes that apply only to parts of the environment. Without looking into the data structure of the environment, we assume that these kinds of changes then only affect the fitness of some possible plans, but do not change the fitness landscape of the domain completely. We thus call a given change function $c$ within a given environment $e \in E$ \emph{reasonable} iff it fulfills the formula:
$$|\{p \in P : |f(p, e) - f(p, c(e))| > \varepsilon \}| \ll |P|.$$

Here, $\varepsilon$ described a small value used as a minimally discernible distance between fitness values. Likewise, the exact meaning of $\ll$ is to be defined by the case. From this definition, it follows that a planner can prepare for a reasonable change by finding a good plan among the plans that are not affected by the reasonable change. When the change occurs, it can then provide a ``quite good'' plan immediately before it even begins to search for good plans among the changed parts of the domain. Thus, to increase planning resilience, we want our planner to not converge on and around the best optimum it has found so far, but to always keep an eye out for other local optima, even when they do not look as promising at the moment.

Note that this behavior can be likened to strategies developed to prevent premature convergence, a problem with metaheuristic search methods that occurs even in static domains~\cite{eiben2003introduction,DBLP:conf/gecco/GaborB17}.

\paraspace{}
\paragraph{Unexpectedness} Even if a planner can prepare for a reasonable change by diversifying, there are often more efficient ways to prepare for expected change: Usually, we would include instances of expected change into the fitness function by simply evaluating the system in the changed environments as well. In that case, the planner can still fully converge on the predicted path of the environment and not spend computational resources on diversification. However, we claim that in most practical applications the future is not completely predictable and changes may happen that the planner cannot anticipate.

We define a change function $c$ to be called \emph{unexpected} iff the planner is not prepared for the change induced, i.e., if the actions it would take in the unchanged environment $e$ differ from the actions it now has to take in the changed environment $c(e)$. Formally, this can be expressed as follows:

$$|\{e \in E : \textit{plan}(c(e)) \not\approx \textit{plan}(e)\}| \gg 0$$

Again, an exact definition of $\gg$ would need to be derived from specific system requirements. Note that this is a purely extrinsic view on unexpectedness. We want to provide a black-box definition of unexpectedness that does not depend on the internal workings of the planner and is thus as general as possible. The intuition behind it is that if there was a way for the planner to know that and how the change $c$ is going to happen when looking at the environment $e$, the plan generated via $\textit{plan}(e)$ would already consider the consequences of said change and thus (to some extent) match the plan for $c(e)$.\footnote{Note that this argument is based on the fact that we defined $\textit{plan}$ in such a way that it tries to optimize for $f(p, e)$ when possible. The result is that we can regard the definitions of ``reasonable'' and ``unexpected'' as upper and lower bounds on the amount of change introduced in the fitness landscape.}

\section{Experiment}
\label{sec:experiment}
To test the validity of our claims about the importance of diversity for planning resilience, we build a model example in which we try to observe the effects of environmental changes as clearly as possible.

\begin{figure}[!t]
\centering\includegraphics[width=0.45\textwidth]{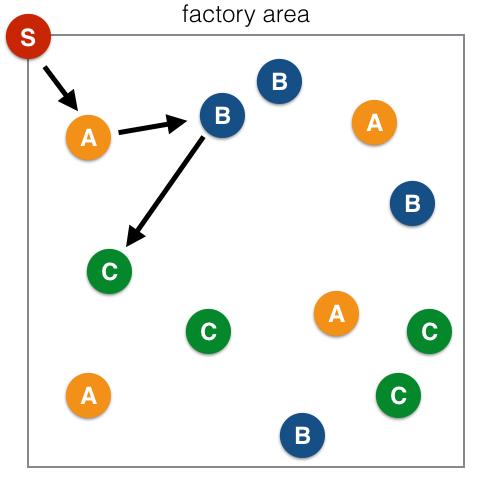}
\caption{Illustration of the factory setup, simplified for only 3 tasks $A, B, C$ and 4 stations for each task. Coming from the starting point $S$, the genetic algorithm needs to determine an as short as possible path that traverses a station for each task in order (see arrows).}
\label{fig:factory}
\end{figure}

\paraspace{}
\paragraph{Scenario} We imagine a smart factory scenario where a work piece carried by a mobile (robotic) agent needs to be processed by a setup of work stations. More specifically, we need to perform the 5 tasks $A, B, C, D, E$ in order on a given work piece as quickly as possible. In order to do so, our factory contains 25 work stations placed randomly on a $500 \times 500$ grid structure. Each work station can only perform one of the tasks, so that our factory has 5 identical work stations to use for any specific task. Given a work piece starting at the top left corner of the grid, we need to determine the shortest route the work piece can travel for it to reach exactly one station of each task in the right order. See Figure~\ref{fig:factory} for a simplified illustration of this setup.

For each run of our experiment, we randomly generate an $n \times m$ matrix $F$ of work station coordinates where each row in $F$ corresponds to a task and each column to an identification number for each of the available work stations for each task. Thus, in our experimental setup we fix $n = 5$ and $m = 5$.

\paraspace{}
\paragraph{Genetic Algorithm} In order to find a short path that fulfills our requirements, we employ a genetic algorithm \cite{eiben2003introduction}. Closely modeling our problem domain, we define the genome as a 5-dimensional vector $v \in \{0, ..., m-1\}^{n}$ so that $v_i$ denotes which of the 5 identical work stations should be visited next in order to fulfill the $i$-th task where $i=0$ denotes the task $A$, $i=1$ denotes task $B$, and so on. The environment provides a mapping from these $v_i$ to their respective positions on the grid, which is used by a distance function $L^E$ for the environment $E$ to compute the traveling distance between two work stations. We then define a function $\textit{waycost}$ to compute the overall length of a given path, summing the Manhattan\footnote{Obviously, real-world mobile transport robots are more likely to navigate in Euclidean space. However, we argue that this is not crucial for the results presented in this paper and choose the computationally simpler approach.} distance $L_1^E$ between all its vertices:

$$\textit{waycost}(v, E) = L_1^E(S, v_0) + \sum_{i=0}^{n-2} L_1^E(v_i, v_{i+1})$$

For the standard genetic algorithm, this $\textit{waycost}$ function is already sufficient as a fitness function $f(v, E) = \textit{waycost}(v, E)$ to evolve a shorter navigation path. It is important to note that while we closely restrict the search space to paths that cross each type of station exactly once (and in the right order), we do \emph{not} aid the genetic algorithm by providing any notion of position in space or the closeness of different stations beyond what is encoded in the $\textit{waycost}$ function above.

For the genome defined above, we use the following evolutionary operators: Mutation chooses a value $i, 0 \leq i < n$ uniformly at random, then generates a new random value $x \in \{0, ..., m-1\}$, assigning $v_i := x$. Recombination happens through uniform crossover on the vectors of two individuals. Furthermore, for all experiments performed in this paper, we use a mutation rate of $0.1$ per individual to provide strong random input and a crossover rate of $0.3$. That means that with a chance of $30\%$ per individual that individual is selected as a first mate for recombination. Two potential mates are then randomly selected from the population: the fitter one is used for as a partner for crossover. We further augment the search by randomly generating some new individuals from scratch each generation. This process (also called hyper-mutation \cite{grefenstette1992genetic}) happens with a chance of $0.1$ per individual in the population.

\paraspace{}
\paragraph{Random Change} The crucial point of this experimental setup is the occurrence of a random change of environmental circumstances. The present experimental setup is fixed to an evaluation time of $100$ generations as earlier experiments have shown our setup of an evolutionary algorithm can easily converge in under $50$ generations. We then define a function for unexpected change $c_A$, which chooses $A$ factory stations at random and effectively disables them. This is implemented by repositioning them to an area far off the usual factory area by adding $(2500, 2500)$ to their respective coordinates. This means that while the plans containing the removed stations are still theoretically feasible and can be assigned a valid $\textit{waycost}$, the increase in $\textit{waycost}$ is so high that no plan containing any of the removed stations should be able to compete with plans contained within the actual factory area when it comes to evolutionary selection.
From a random initial factory layout $F$ we generate two changed factory layouts $F_1 = c_A(F), F_2 = c_A(F)$ by applying the randomized change function $c_A$. Because we want to be able to compare the scale of fitness values before and after the unexpected change more easily, we start the evolutionary algorithm on the factory configuration $F_1$ that is already ``missing'' a few stations. After $50$ generations, we switch to factory configuration $F_2$, which has $A$ stations disabled as well, but probably different ones.\footnote{It is important to note that this setup means that in many cases none of the stations that go bad during the switch are even included in the best path found by the genetic algorithm. In these cases, the evolutionary process does not have to adapt in any way. In order to analyze the cases when the removal of stations actually does make a huge difference, we need to execute the experiment multiple times. We chose this approach because it allows us use an unbiased change function as opposed to a change function that specifically targets the workstations actually used throughout the experiment. The realm of biased, even directly adversarial change functions is an interesting topic of future research.} 

Note that this change is reasonable for small $A$ (according to the definition above) because it only affects the fitness of a maximum of $2*A$ possible plans, i.e., those plans which include at least one of the ``wrong'' machines in $F_1$ or $F_2$. Furthermore, the change is unexpected as the shakeup of the stations' positioning is communicated to the evolutionary algorithm only via the change of the $\textit{waycost}$ function's values in its fitness evaluation step and thus leaves the adaptation process without any chance of anticipating that event. Nonetheless, the individuals of the evolutionary process are constantly evaluated according to their fitness in the current state of affairs, thus forcing them to adapt to the new situation in order to keep up once reached levels of fitness values.

\section{Approach}
\label{sec:approach}
We attempt to solve the problem described above using evolutionary algorithms. Evolutionary algorithms have already been applied successfully to many instances of online adaptation, i.e., problems with a changing fitness function \cite{nikolos2003evolutionary,perez2013rolling,gaina2017rolling}. They are an instance of metaheuristic search algorithms and work by emulating natural evolution.

\paraspace{}
\paragraph{Diversity in Genetic Algorithms} In the standard scenario, once the fitness function changes, previously good solutions can possibly be evaluated to have very bad fitness and are thus removed from the evolutionary process. However, if the genetic search has already converged to a local optimum, it can be very hard for the search process to break out of it, because when all known solutions lie very closely together in the solution space, there is no clear path along which the population must travel in order to improve. The problem of a genetic search getting stuck in a local optimum with little chance to reach the global optimum (or at least much better local ones) is called \emph{premature convergence} \cite{eiben2003introduction}. It is known that the diversity among the members in the population has a strong impact on the evolutionary process's likelihood to converge too early. The Diversity-Guided Evolutionary Algorithm (DGEA) observes a population's diversity throughout the evolutionary process and takes action when it falls below a given threshold \cite{ursem2002diversity}.

For online genetic algorithms, we show that maintaining a certain level of diversity throughout the population helps to react better to the change occurring in the environment. To this end, we apply two possible measurements for diversity, which we will both test for the above scenario. In either case, we transform the genetic algorithm's fitness function to a \emph{multi-objective optimization} problem \cite{konak2006multi,DBLP:conf/gecco/GaborB17,gabor2018inheritance} with a weighting parameter $\lambda$, yielding a fitness function $f$ depending on the individual to be evaluated $v$, the environment $E$, and the population $P$ as a whole:

$$f(v, E, P) = \textit{waycost}(v, E) + \lambda * \textit{similaritycost}(v, P)$$

It is important to note that in order to meaningfully define the diversity of one individual, we need to compare it to the rest of the population, causing us to introduce the population $P$ as an additional parameter to the fitness function.\footnote{In general, we might want approximate this comparison by using a sample drawn from the population or another estimate instead. Likewise, we could consider computing diversity not only against the current generation of individuals but also against a selection of individuals from the past, using for example a ``hall of fame'' approach \cite{rosin1997new}. The evaluation of such techniques is left for future research.} The fitness function thus becomes a relative measure with respect to other individuals in the population. This makes it necessary to re-evaluate fitness in each generation even for unchanged individuals. However, since we assume changes in the environment and thus the fitness function may occur during the online execution of the genetic algorithm anyway, this model seems to fit our situation. We can now define two different diversity measures by providing a definition for the $\textit{similaritycost}$ function, which penalizes low diversity.

\paraspace{}
\paragraph{Domain-Distance Diversity} This can be thought of as the more standard approach to diversity in search and optimization problems. In fact, the authors of \cite{wineberg2003underlying} show that many common diversity measurements are quite similar to this basic method: We define a simple distance measure between the individuals in the solution space. For a discrete, categorial problem like the one presented here, there is little alternative to just counting the specific differences in some way.

$$\textit{similaritycost}_{dom}(v, P) = -n + \sum_{i=0}^{n-1} \sum_{j=0}^{|P|} C(v_i, P(j)_i)$$

$$\textit{where } C(x, y) = \begin{cases} 
      1 & \textit{if } \;\; x = y\\
      0 & \textit{otherwise}
   \end{cases}$$
   
Note that we write $P(j)$ to access the $j$-th individual of the population and $|P|$ to represent the amount of individuals in a population. We subtract $n$ from the sum because the given individual $v \in P$ is still part of the population and thus adds a cost of $n$ by matching itself perfectly. We thus maintain the (cosmetic) property that in a population of completely different individuals, the average similarity is $0$.

While the implementation of this diversity measure looks pretty straightforward, it requires complete prior knowledge of the search space provided and and thus introduces further dependencies. For example, the above definition is unfit for continuous search spaces and while a continuous $\textit{similaritycost}$ function may easily be thought up, optimization problems consisting of a mix of discrete and continuous variables then require more weighting parameters to adequately combine the scales over which the respective $\textit{similaritycost}$ functions operate.

\paraspace{}
\paragraph{Genealogical Diversity} As a more different comparison we implemented a inheritance-based diversity estimate introduced in \cite{DBLP:conf/gecco/GaborB17}. The aim of genealogical diversity is to utilize those parts of the domain knowledge that are already encoded in the setup of the genetic algorithm, i.e., the mutation and recombination function the human developer is required to code for the specific genome anyway. We can thus try to quantify the difference between two individuals by estimating the amount of evolution steps it took to develop these different instances of solution candidates. This yields a measure of ``relatedness'' between individuals not unlike genealogical trees in biology or human ancestry. If all individuals in a population are closely related (sibling or cousins, e.g.), we know that there can only be limited genetic difference between them and thus estimate a low diversity for the respective individuals with respect to that population.

However, instead of building and traversing a genealogical tree, the implementation of genealogical diversity used in \cite{DBLP:conf/gecco/GaborB17} employs a technique inspired by the way genetic genealogical trees are constructed from the analysis from genomes in biological applications: For this approach, we first need to augment the individuals' genome by a series of $t$ trash bits $b_k \in \{0, 1\}, k \in \mathbb{N}, 0 \leq k < t$. For our experiment, $t = 16$ has proven to be reasonable. However, we do not change the $\textit{waycost}$ fitness function, so that it does not recognize the additional data added to the genome. This leads to the trash bits not being subjected to selection pressure from the primary objective of the genetic algorithm.

As the trash bits are randomly initialized like the other variables in the genome, every individual of the first generation should most probably start out with a very different trash bitstring from anyone else's, given that we choose the length of the trash bitstring sufficiently large. Without direct selection pressure, there is no incentive for individuals to adapt their trash bitstring in any specific way. However, the trash bits are still subjected to mutation and recombination, i.e., whenever a specific individual is chosen for mutation, a random mutation is performed on the trash bitstring as well and whenever a recombination operation is executed for two individuals, their trash bitstrings are likewise recombined. In our implementation at hand, we use one-bit flip for mutation and uniform crossover for recombination.

Using the definition of a comparison function $C$ as provided above, we can thus define the $\textit{similaritycost}$ function for genealogical diversity as follows:

$$\textit{similaritycost}_{gen}(v, P) = -t + \sum_{i=0}^{t-1} \sum_{j=0}^{|P|} C(v_{n+i}, P(j)_{n+i})$$

Again, we subtract $t$ to ignore self-similarity when iterating over the population. It should be noted that when accessing the $(n+i)$-th component of an individual inside the sum, we are protruding into the dimensions solely populated by trash bits, retrieving the $i$-th trash bit of said individual.

In order to compute the similarity between two individuals, we now only consider the trash bits, for which we always have the same distance metric regardless of the actual problem domain of the original genetic algorithm. Domain logic is only used indirectly, as the measure we estimate can be regarded as the edit distance between two individuals using the genetic operators the evolutionary process is equipped with. However, since the trash bits are inherited by individuals from their parents and without direct selection pressure, they are not biased toward values resulting in higher fitness; yet, they are still a sufficient representation of the genealogy of an individual, as we show in the following section.

\section{Results}
\label{sec:results}

\begin{figure}[!t]
\centering\includegraphics[width=0.485\textwidth]{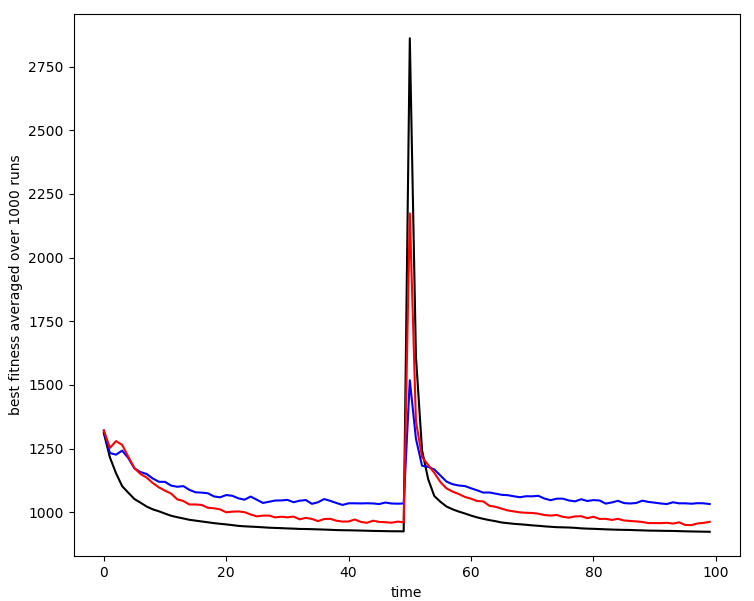}
\caption{Best (i.e., lowest-valued) fitness for current generation averaged over 1000 runs. While the evolutionary algorithm without any recognition of diversity (black) shows a steep spike at the time of the environmental change (after 50 generations), genealogically (red) and the domain-dependent (blue) diverse genetic algorithms manage to mitigate the negative amplitude to varying extent.}
\label{fig:typical}
\end{figure}

\begin{figure}[!t]
\centering\includegraphics[width=0.485\textwidth]{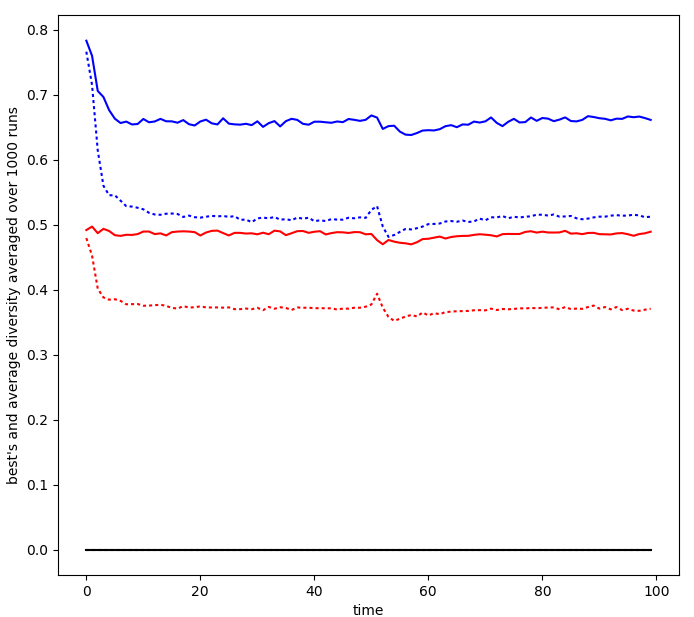}
\caption{Diversity measures of the top individual (solid line) as well as the population average diversity (dotted line) per generation averaged over 1000 runs. We draw both genealogical (red) and domain-dependent (blue) diversity into the same figure as they are both normalized on $[0; 1]$, even though no direct translation is possible between their values. In both cases, the population's average diversity shows a specific behavior following the unexpected change.}
\label{fig:typical-div}
\end{figure}

In order to evaluate the benefit of the presented approaches, we simulate the different behavior of genetic algorithms when using the presented diversity measures or no diversity measure at all. In order to achieve a meaningful result considering the highly probabilistic nature of the applied method to generate scenarios, we perform the evaluation on 1000 different scenarios. Figure~\ref{fig:typical} shows the top fitness achieved at a specific point in time by a single run averaged over all 1000 runs. By taking a look at the optimization process as a whole, it can be seen that a great deal of improvement compared to the random initialization is done during the first steps of evolution, giving an estimate of how good the achieved solutions are in relation to ``just guessing''. In Figure~\ref{fig:typical-div} we show the respective diversity measurements from these runs. 

We can observe that the diversity-aware algorithms show a slower learning rate in the beginning, since they do not only optimize the plotted primary fitness function, but also the diversity function and thus cannot focus as well on better primary results. However, once the environmental change occurs, they are likewise better prepared for a change in fitness and react with a much smaller increase in $\textit{waycost}$ than the standard genetic algorithm. In a scenario like ours, where a smart factory needs to be able to efficiently dispatch new workpieces at all times, this can be a huge advantage. We observe that following the unexpected change, average diversity first increases as well-established ``families'' of similar individuals die out. Due to a new convergence process, diversity then drops until good solutions are found. Finally, diversity seems to reach a similar level as before the unexpected change. The ``right'' amount of diversity is naturally controlled by the parameter $\lambda$ of the combined fitness function. For these experiments we found parameters $\lambda=1500$ for domain-dependent diversity and $\lambda=2500$ for genealogical diversity via systematic search.

The definition of ``right'', however, depends on the problem domain. In most practical cases, we expect some (non-functional) requirements to be present, which specify the robustness properties we want to uphold. For now, these properties must then be verified via statistic testing. Deriving (statistical or hard) guarantees from a stochastic search process like an evolutionary algorithm is still an interesting topics of future work. Goven no further requirements for consistent quality of service, a reasonable setting for $\lambda$ might achieve that the online planner does not perform worse than a random planner at any point in time, even at the moment of unexpected change.

\begin{figure}[!t]
\centering\includegraphics[width=0.46\textwidth]{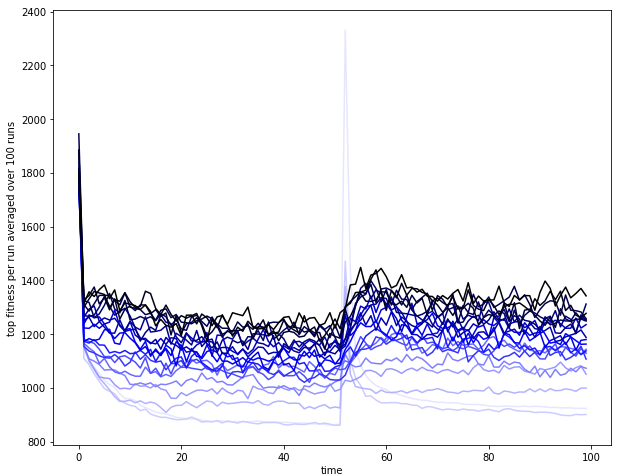}
\caption{Top fitness for current generation averaged over only 100 runs each, plotted for $\lambda = 500 * z, z \in \mathbb{N}, 0 \leq z < 20$ using domain-dependent diversity. The darker the color of the line, the higher is the depicted $\lambda$ value.}
\label{fig:lame}
\end{figure}

\begin{figure}[!t]
\centering\includegraphics[width=0.46\textwidth]{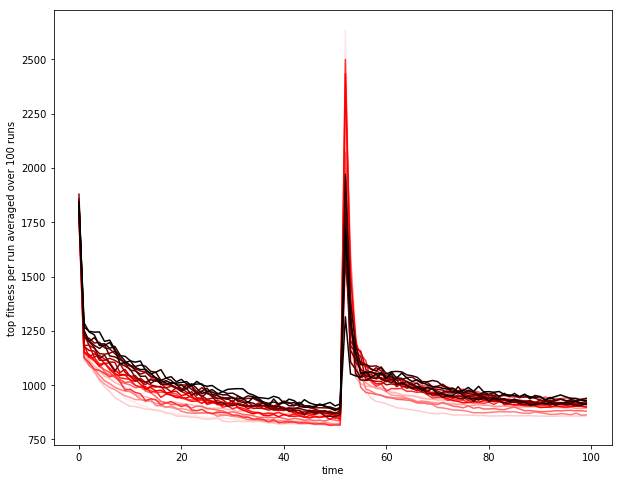}
\caption{Top fitness for current generation averaged over only 100 runs each, plotted for $\lambda = 500 * z, z \in \mathbb{N}, 0 \leq z < 20$ using genealogical diversity. The darker the color of the line, the higher is the depicted $\lambda$ value.}
\label{fig:cool}
\end{figure}

Figures~\ref{fig:lame} and~\ref{fig:cool} show that systematic search, including the random population's value before the evolutionary process starts: the fitness achieved by the domain-dependent and the genealogical genetic algorithm, respectively, strongly depends on the choice of parameter $\lambda$, i.e., how to distribute focus between the primary objective (small $\textit{waycost}$) and the secondary objective (high diversity). Experiments have shown, that diversity-aware genetic algorithms can show a variety of behaviors for different $\lambda$. To provide an intuition about the effects various settings for $\lambda$ have on the algorithm's performance, we can see that higher values of $\lambda$ generally cause the evolutionary search to produce less optimal results but to perform more stable when facing unexpected change. For the domain-dependent diversity, this phenomenon shows stronger with higher $\lambda$-values showing almost no impact of the unexpected change but relatively bad results in general. The approach of genealogical diversity seems to be a bit more robust to the setting of $\lambda$ in that it still clearly shows a tendency to optimize over time.

\begin{figure}[!t]
\centering\includegraphics[width=0.46\textwidth]{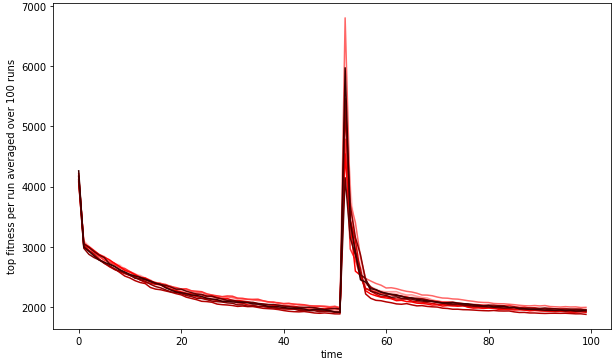}
\caption{Top fitness for current generation averaged over 100 runs each, plotted for $t = 2^z, z \in \mathbb{N}, 0 \leq z < 10$ using genealogical diversity. The darker the color of the line, the higher is the depicted $t$ value.}
\label{fig:tau}
\end{figure}

We chose to showcase genealogical diversity specifically because it works on a rather domain-independent level and introduces only few parameters. Furthermore, it is rather robust with respect to the choice of said parameters. For the length of the used bitstring $t$, Figure~\ref{fig:tau} shows that on all but the smallest values for $t$ the genetic algorithm performs most similarly. Especially rather large values for $t$ (that still take up very little memory) do not show any deterioration in the planner's behavior, which means that the choice for that parameter can be made rather comfortably.

\begin{figure}[!t]
\centering\includegraphics[width=0.5\textwidth]{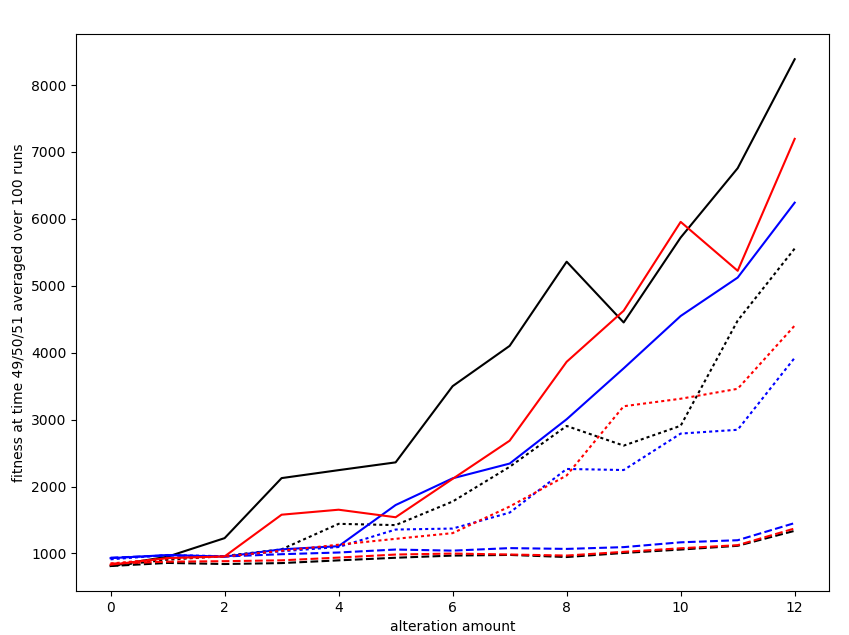}
\caption{Analysis of the fitness amplitude around unexpected change of varying intensity for the non-diverse (black), the genealogically diverse (red) and the domain-dependent diverse (blue) evolutionary algorithm respectively, plotted against the parameter $A$ for the alternation amount of the change function $c_A$, all results averaged over 100 runs. The dashed line shows the population's top fitness value just before the change (generation 49). The solid line shows the top fitness just at the moment of unexpected change (generation 50). The dotted line shows the fitness one generation later (generation 51), when it has started to improve again.}
\label{fig:alteration}
\end{figure}

We also analyze \emph{how much} change a diversity-aware planner can handle. Figure~\ref{fig:alteration} shows the behavior of the three exemplary planners just around the moment of unexpected change for various amounts of change they are subjected to. Naturally, bigger (and thus un-reasonable) change can impact even diverse system. The increase in costs for the large alterations in the generation-49-line (dashed) shows that on the upper end of the scale we started generating problem instances that generally have fewer good solutions. For more reasonable change ($A \leq 8$, which still means that up to $16$ out of $25$ machine positions may be changed), both diversity-aware algorithms perform comparably and clearly better than the non-diverse planner. Most remarkably, the domain-dependent variant manages to cope with changes $A \leq 4$ with almost no consequence for its performance.

\section{Related Work}
\label{sec:rel}

The notion of diversity is researched in many different fields of science. In this Section, we discuss a few of these and their relation to the issue presented in this paper. To the author's knowledge, the issue of planning resilience is a rather novel one and applying genetic diversity to promote it is a unique contribution of this paper.

\paraspace{}
\paragraph{Planning and Reinforcement Learning} Rolling horizon genetic algorithms for online planning are widely used in real-time general video game playing \cite{perez2013rolling,gaina2017rolling}. However, these approaches typically optimize with respect to the expectation of reward, thus they suffer from the drawbacks of non-diverse planning as discussed in this work. Statistical online planning has recently attracted a fair amount of research interest \cite{weinstein2013open,eastwood2016safe,belzner2016time}, also due to the successful combination with deep learning technology \cite{silver2016mastering,silver2017mastering,DBLP:journals/corr/AnthonyTB17}. In general, diversity as a consideration for resilient planning is orthogonal to these approaches and could straightforwardly be combined with recent developments from the online planning literature.

We also see a relation to another recent line of research in reinforcement learning that explores ways of modeling aleatoric or epistemic uncertainty about future rewards or action selection mechanisms as distributions rather than by expectation \cite{bellemare2017distributional,dabney2017distributional,DBLP:journals/corr/HaarnojaTAL17,schulman2017equivalence,ghavamzadeh2015bayesian,o2017uncertainty}. This enables learning and decision making agents to explicitly deal with multimodal distributions of utility. It allows to incorporate risk and uncertainty into the decision making process, and to effectively deal with the exploration-exploitation tradeoff. In particular, distributional approaches foster the learning of diverse behavior, yielding robust transfer of learned skills to new, unseen situations \cite{DBLP:journals/corr/HaarnojaTAL17}.

\paraspace
\paragraph{Diversity in Software} In software engineering, diversity often takes the form of generating, offering, or using functionally equivalent variants of software artifacts during software development or deployment \cite{schaefer2012software}. An extensive survey of current techniques is given in \cite{konak2006multi}. However, all of these differ from the approach in this paper in that we explicitly search for \emph{functional} alternatives in the context of this paper, i.e., we want our diverse solutions to represent solutions to different problems (in order to possibly anticipate future problems) and not different solutions of equal quality to the same problem.

Still, the techniques presented in literature to exploit the prevalence of multiple instances of the same software artifact during runtime might be applied to variants generated by a diverse genetic algorithm as well. The work in this paper can be regarded as a first step to expose the population-based view of diversity within an automated search process with the process of software development. Similar trends in software engineering are discussed in \cite{wirsing2015software, holzl2015continuous}.

\paraspace{}
\paragraph{Diversity in Genetic Algorithms} Genetic algorithms make up a vast field of research. Regarding the basic definitions, this work follows the comprehensive description in \cite{eiben2003introduction}. Diversity has been recognized as an important indicator for good performance, although mainly applied to the static scenarios of offline adaptation: The authors of \cite{squillero2016divergence} provide an extensive survey of various methods to enforce diversity in genetic algorithms. These fall into the categories of external methods controlling the evolutionary process ``from the outside'' \cite{deb2000fast,ursem2002diversity} and methods integrating diversity as an additional objective into the genetic algorithm, using the concepts of multi-objective genetic optimization \cite{laumanns2002combining,segurax2016using}. Following the biological inspiration, the aptitude of genetic algorithms to an online setting with a changing environment has been thoroughly analyzed \cite{vavak1996comparison,bredeche2009line}.

The author of \cite{grefenstette1992genetic} describes a problem setting not unlike the one presented in this paper, i.e., the combination of maintaining diversity and searching in a changing environment. The issue of premature convergence is tackled by integrating a certain amount of random search into the genetic algorithm by performing hyper-mutation. This has since become standard procedure and is included in all genetic algorithms presented in this paper, which aims to further improve the resilience of the search process.

Most recently, the authors of \cite{kinneer2018managing} tackled the issue of using an evolutionary algorithm as an online planner for a complex software system. While they discuss high diversity as a key factor in achieving better re-planning results, they use diversity purely as an observation not as a direct goal of the evolutionary process. Instead, they too resort to an operator akin to high amounts of hyper-mutation to increase diversity by creating a new population that only inherits certain parts of the old population. To this end, the system must be able to directly recognize the event of an unexpected change after it has happened.

It is important to note that a lot of literature about diversity in genetic algorithms (or metaheuristic search in general) is concerned about covering the frontier of Pareto-optimal solutions in the search space \cite{horn1994niched}. The notion of diversity used in this paper, however, is a more genetic one and has as one of its main features that is \emph{not} automatically derived from the nature of the fitness function. Interesting connections to game theory may still be made but are outside the scope of this work.

\paraspace{}
\paragraph{Resilience and Robustness} The preparation for unexpected or previously wrongly modeled change is an important issue for the practical application of machine learning in industry \cite{DBLP:journals/corr/AmodeiOSCSM16}. From an engineer's point of view, the diversity of the population of plans can be regarded as a typical \emph{non-functional requirement (NFR)} with the cost of the plan representing the functional requirement. Applying NFR engineering processes to self-adaptive systems is still a new idea and a clear canon of relevant NFRs for these new challenges has not yet been found \cite{de2013software,belzner2016software}.

\section{Conclusion}
\label{sec:conclusion}

Since we expect future software systems to be increasingly self-adaptive and self-managing, we can also expect them to feature one or multiple components tasked with online planning. Online planning allows systems to learn to optimize their behavior in the face of a moving target fitness. However, it comes with a few pitfalls, one of which is the fact even small changes in the target fitness can have detrimental effects on the current plans' performance. It is thus imperative to keep an eye on a healthy level of diversity in our pool of alternative plans. As we have shown, this can severely soften the blow to overall performance, should only a few plans become impractical due to external circumstances.\footnote{It still holds that if we allow arbitrary changes in the environment, it is always possible to design a completely new fitness function so that any given instance of an evolutionary process becomes arbitrarily bad with respect to the new altered fitness function. This is due to the No-Free-Lunch theorem \cite{wolpert1997no}. For realistic scenarios, however, there usually is a limit to how quickly and how drastically the fitness function is expected to change. A thorough analysis of those limits for some practical domains may present an interesting point for further research.}


The diversity of a planner functions as a non-functional requirement for classic applications. Certain levels of desired diversity may be specified in order to augment system architectures that revolve around the optimization process of the system in order to provide flexibility on the component level \cite{holzl2016continuous}. This should be expected to strongly influence other properties commonly applied to complex self-adaptive systems like robustness or flexibility.

On an application level, the introduced concept of diversity-aware optimization may prove especially useful when the reduction in amplitude of fitness causes the system behavior to fall below a predefined quality threshold (or to do so more often at least). A diversity-aware planner might then be able to continue working as usual as its back-up plans fulfill the required quality agreement just as well while a non-diverse planner might more often feel the need to stop the execution of its plans (and thus halt the system in general) until it reaches a new plan of acceptable quality. In this case, we may formulate a non-functional requirement such as planning resilience, measuring how frequent and how big unexpected changes need to be in order to push the planner out of its quality requirements. Using the parameter $\lambda$, engineers can adjust the focus point of the planning component between performance and resilience optimization. How well statistical judgements can be made about said resilience property still needs to be evaluated, though.

It is up to future research to determine how the concept of diversity (especially genealogical diversity) generalizes for other optimization techniques like the cross-entropy method or simulated annealing. One way to integrate these techniques into the framework defined in this paper may be to set up a pool of solution candidates via ensemble learning \cite{hart2018evolution}.

Embracing diversity seems especially promising in search-based software testing (SBST) as test suites need to adapt faster to new possible exploits. In DevOps, developers push relatively small updates that need testing more frequently. Nonetheless, the changes applied to the code by the developer usually fall into the category of unexpected change as we defined it in this paper. That means, that diverse test generators could possible adapt quicker to the new software system under test. The mutual influence between diversity-aware evolutionary algorithms and co-evolutionary approaches\footnote{For example, when SBST is used to analyze a system under test that is by itself capable of adapting and evolving, the complete testing cycle features two adversary evolutionary algorithms and is thus considered \emph{co-evolutionary} \cite{harman2012search}.} may be an interesting point of further research \cite{rosin1997new}. A likewise connection in biological systems has been found \cite{berenos2010antagonistic}.

Many of the theoretical foundations explaining the ideal structure of a population for various optimization purposes are still unexplored. For instance, we assumed an unpredictable but neither explicitly hostile nor cooperative environment. Any scenario where the change occurs not only unexpected but intentional is likely to have fundamentally different properties. 

We focused our study on the implications of using diversity within a planner and how the resilience to environmental change may be indicated in a quantifiable way. We have shown that diversity during planning can aid planning resilience in the face of change. Furthermore, we can employ such method in a domain-independent way using genealogical diversity and still achieve valuable results. Software engineering frameworks and processes are now needed to expose desired NFRs like planning resilience to the software and system design and test them adequately.



\bibliographystyle{IEEEtran}
\bibliography{references}

\begin{thebibliography}{10}
\providecommand{\url}[1]{#1}
\csname url@samestyle\endcsname
\providecommand{\newblock}{\relax}
\providecommand{\bibinfo}[2]{#2}
\providecommand{\BIBentrySTDinterwordspacing}{\spaceskip=0pt\relax}
\providecommand{\BIBentryALTinterwordstretchfactor}{4}
\providecommand{\BIBentryALTinterwordspacing}{\spaceskip=\fontdimen2\font plus
\BIBentryALTinterwordstretchfactor\fontdimen3\font minus
  \fontdimen4\font\relax}
\providecommand{\BIBforeignlanguage}[2]{{%
\expandafter\ifx\csname l@#1\endcsname\relax
\typeout{** WARNING: IEEEtran.bst: No hyphenation pattern has been}%
\typeout{** loaded for the language `#1'. Using the pattern for}%
\typeout{** the default language instead.}%
\else
\language=\csname l@#1\endcsname
\fi
#2}}
\providecommand{\BIBdecl}{\relax}
\BIBdecl

\bibitem{harman2012search}
M.~Harman, S.~A. Mansouri, and Y.~Zhang, ``Search-based software engineering:
  Trends, techniques and applications,'' \emph{ACM Computing Surveys (CSUR)},
  vol.~45, no.~1, p.~11, 2012.

\bibitem{de2013software}
R.~De~Lemos, H.~Giese, H.~A. M{\"u}ller, M.~Shaw, J.~Andersson, M.~Litoiu,
  B.~Schmerl, G.~Tamura, N.~M. Villegas, T.~Vogel \emph{et~al.}, ``Software
  engineering for self-adaptive systems: A second research roadmap,'' in
  \emph{Software Engineering for Self-Adaptive Systems II}.\hskip 1em plus
  0.5em minus 0.4em\relax Springer, 2013.

\bibitem{wirsing2015software}
M.~Wirsing, M.~H{\"o}lzl, N.~Koch, and P.~Mayer, \emph{Software Engineering for
  Collective Autonomic Systems: The ASCENS Approach}.\hskip 1em plus 0.5em
  minus 0.4em\relax Springer, 2015.

\bibitem{DBLP:journals/corr/AmodeiOSCSM16}
\BIBentryALTinterwordspacing
D.~Amodei, C.~Olah, J.~Steinhardt, P.~Christiano, J.~Schulman, and
  D.~Man{\'{e}}, ``Concrete problems in {AI} safety,'' \emph{CoRR}, vol.
  abs/1606.06565, 2016. [Online]. Available:
  \url{http://arxiv.org/abs/1606.06565}
\BIBentrySTDinterwordspacing

\bibitem{thoben2017industrie}
K.-D. Thoben, S.~Wiesner, and T.~Wuest, ``“industrie 4.0” and smart
  manufacturing--a review of research issues and application examples,''
  \emph{Int. J. of Automation Technology Vol}, vol.~11, no.~1, 2017.

\bibitem{kephart2003vision}
J.~O. Kephart and D.~M. Chess, ``The vision of autonomic computing,''
  \emph{Computer}, vol.~36, no.~1, pp. 41--50, 2003.

\bibitem{chen2015requirements}
B.~Chen, X.~Peng, Y.~Yu, and W.~Zhao, ``Requirements-driven self-optimization
  of composite services using feedback control,'' \emph{IEEE Transactions on
  Services Computing}, vol.~8, no.~1, pp. 107--120, 2015.

\bibitem{arcaini2015modeling}
P.~Arcaini, E.~Riccobene, and P.~Scandurra, ``Modeling and analyzing mape-k
  feedback loops for self-adaptation,'' in \emph{Proceedings of the 10th
  International Symposium on Software Engineering for Adaptive and
  Self-Managing Systems}.\hskip 1em plus 0.5em minus 0.4em\relax IEEE Press,
  2015, pp. 13--23.

\bibitem{belzner2016software}
L.~Belzner, M.~T. Beck, T.~Gabor, H.~Roelle, and H.~Sauer, ``Software
  engineering for distributed autonomous real-time systems,'' in
  \emph{Proceedings of the 2nd International Workshop on Software Engineering
  for Smart Cyber-Physical Systems}.\hskip 1em plus 0.5em minus 0.4em\relax
  ACM, 2016, pp. 54--57.

\bibitem{bankes2010robustness}
S.~C. Bankes, ``Robustness, adaptivity, and resiliency analysis.'' in
  \emph{AAAI fall symposium: complex adaptive systems}, vol.~10, 2010.

\bibitem{florio2013constituent}
V.~D. Florio, ``On the constituent attributes of software and organizational
  resilience,'' \emph{Interdisciplinary Science Reviews}, vol.~38, no.~2, 2013.

\bibitem{eiben2003introduction}
A.~E. Eiben and J.~E. Smith, \emph{Introduction to evolutionary
  computing}.\hskip 1em plus 0.5em minus 0.4em\relax Springer, 2003, vol.~53.

\bibitem{DBLP:conf/gecco/GaborB17}
T.~Gabor and L.~Belzner, ``Genealogical distance as a diversity estimate in
  evolutionary algorithms,'' in \emph{Genetic and Evolutionary Computation
  Conference, Berlin, Germany, July 15-19, 2017, Companion Material
  Proceedings}, P.~A.~N. Bosman, Ed.\hskip 1em plus 0.5em minus 0.4em\relax
  {ACM}, 2017.

\bibitem{grefenstette1992genetic}
J.~J. Grefenstette \emph{et~al.}, ``Genetic algorithms for changing
  environments,'' in \emph{PPSN}, vol.~2, 1992, pp. 137--144.

\bibitem{nikolos2003evolutionary}
I.~K. Nikolos, K.~P. Valavanis, N.~C. Tsourveloudis, and A.~N. Kostaras,
  ``Evolutionary algorithm based offline/online path planner for uav
  navigation,'' \emph{IEEE Transactions on Systems, Man, and Cybernetics, Part
  B (Cybernetics)}, vol.~33, no.~6, pp. 898--912, 2003.

\bibitem{perez2013rolling}
D.~Perez, S.~Samothrakis, S.~Lucas, and P.~Rohlfshagen, ``Rolling horizon
  evolution versus tree search for navigation in single-player real-time
  games,'' in \emph{Proceedings of the 15th annual conference on Genetic and
  evolutionary computation}.\hskip 1em plus 0.5em minus 0.4em\relax ACM, 2013,
  pp. 351--358.

\bibitem{gaina2017rolling}
R.~D. Gaina, S.~M. Lucas, and D.~Perez-Liebana, ``Rolling horizon evolution
  enhancements in general video game playing,'' in \emph{Computational
  Intelligence and Games (CIG), 2017 IEEE Conference on}.\hskip 1em plus 0.5em
  minus 0.4em\relax IEEE, 2017.

\bibitem{ursem2002diversity}
R.~K. Ursem, ``Diversity-guided evolutionary algorithms,'' in \emph{Internat.
  Conference on Parallel Problem Solving from Nature}.\hskip 1em plus 0.5em
  minus 0.4em\relax Springer, 2002.

\bibitem{konak2006multi}
A.~Konak, D.~W. Coit, and A.~E. Smith, ``Multi-objective optimization using
  genetic algorithms: A tutorial,'' \emph{Reliability Engineering \& System
  Safety}, vol.~91, no.~9, pp. 992--1007, 2006.

\bibitem{gabor2018inheritance}
T.~Gabor, L.~Belzner, and C.~Linnhoff-Popien, ``Inheritance-based diversity
  measures for explicit convergence control in evolutionary algorithms,'' in
  \emph{The Genetic and Evolutionary Computation Conference (GECCO)}.\hskip 1em
  plus 0.5em minus 0.4em\relax ACM, 2018.

\bibitem{rosin1997new}
C.~D. Rosin and R.~K. Belew, ``New methods for competitive coevolution,''
  \emph{Evolutionary Computation}, vol.~5, no.~1, pp. 1--29, 1997.

\bibitem{wineberg2003underlying}
M.~Wineberg and F.~Oppacher, ``The underlying similarity of diversity measures
  used in evolutionary computation,'' in \emph{Genetic and Evolutionary
  Computation (GECCO 2003)}.\hskip 1em plus 0.5em minus 0.4em\relax Springer,
  2003, pp. 206--206.

\bibitem{weinstein2013open}
A.~Weinstein and M.~L. Littman, ``Open-loop planning in large-scale stochastic
  domains.'' in \emph{AAAI}, 2013.

\bibitem{eastwood2016safe}
R.~Eastwood, R.~Alexander, and T.~Kelly, ``Safe multi-objective planning with a
  posteriori preferences,'' in \emph{High Assurance Systems Engineering (HASE),
  2016 IEEE 17th International Symposium on}.\hskip 1em plus 0.5em minus
  0.4em\relax IEEE, 2016.

\bibitem{belzner2016time}
L.~Belzner, ``Time-adaptive cross entropy planning,'' in \emph{Proceedings of
  the 31st Annual ACM Symposium on Applied Computing}.\hskip 1em plus 0.5em
  minus 0.4em\relax ACM, 2016.

\bibitem{silver2016mastering}
D.~Silver, A.~Huang, C.~J. Maddison, A.~Guez, L.~Sifre, G.~Van Den~Driessche,
  J.~Schrittwieser, I.~Antonoglou, V.~Panneershelvam, M.~Lanctot \emph{et~al.},
  ``Mastering the game of go with deep neural networks and tree search,''
  \emph{Nature}, vol. 529, no. 7587, 2016.

\bibitem{silver2017mastering}
D.~Silver, J.~Schrittwieser, K.~Simonyan, I.~Antonoglou, A.~Huang, A.~Guez,
  T.~Hubert, L.~Baker, M.~Lai, A.~Bolton \emph{et~al.}, ``Mastering the game of
  go without human knowledge,'' \emph{Nature}, vol. 550, no. 7676, 2017.

\bibitem{DBLP:journals/corr/AnthonyTB17}
\BIBentryALTinterwordspacing
T.~Anthony, Z.~Tian, and D.~Barber, ``Thinking fast and slow with deep learning
  and tree search,'' \emph{CoRR}, vol. abs/1705.08439, 2017. [Online].
  Available: \url{http://arxiv.org/abs/1705.08439}
\BIBentrySTDinterwordspacing

\bibitem{bellemare2017distributional}
M.~G. Bellemare, W.~Dabney, and R.~Munos, ``A distributional perspective on
  reinforcement learning,'' \emph{arXiv preprint arXiv:1707.06887}, 2017.

\bibitem{dabney2017distributional}
W.~Dabney, M.~Rowland, M.~G. Bellemare, and R.~Munos, ``Distributional
  reinforcement learning with quantile regression,'' \emph{arXiv preprint
  arXiv:1710.10044}, 2017.

\bibitem{DBLP:journals/corr/HaarnojaTAL17}
\BIBentryALTinterwordspacing
T.~Haarnoja, H.~Tang, P.~Abbeel, and S.~Levine, ``Reinforcement learning with
  deep energy-based policies,'' \emph{CoRR}, vol. abs/1702.08165, 2017.
  [Online]. Available: \url{http://arxiv.org/abs/1702.08165}
\BIBentrySTDinterwordspacing

\bibitem{schulman2017equivalence}
J.~Schulman, P.~Abbeel, and X.~Chen, ``Equivalence between policy gradients and
  soft q-learning,'' \emph{arXiv preprint arXiv:1704.06440}, 2017.

\bibitem{ghavamzadeh2015bayesian}
M.~Ghavamzadeh, S.~Mannor, J.~Pineau, A.~Tamar \emph{et~al.}, ``Bayesian
  reinforcement learning: A survey,'' \emph{Foundations and
  Trends{\textregistered} in Machine Learning}, vol.~8, no. 5-6, pp. 359--483,
  2015.

\bibitem{o2017uncertainty}
B.~O'Donoghue, I.~Osband, R.~Munos, and V.~Mnih, ``The uncertainty bellman
  equation and exploration,'' \emph{arXiv:1709.05380 preprint}, 2017.

\bibitem{schaefer2012software}
I.~Schaefer, R.~Rabiser, D.~Clarke, L.~Bettini, D.~Benavides, G.~Botterweck,
  A.~Pathak, S.~Trujillo, and K.~Villela, ``Software diversity: state of the
  art and perspectives,'' 2012.

\bibitem{holzl2015continuous}
M.~H{\"o}lzl and T.~Gabor, ``Continuous collaboration: a case study on the
  development of an adaptive cyber-physical system,'' in \emph{Proceedings of
  the First International Workshop on Software Engineering for Smart
  Cyber-Physical Systems}.\hskip 1em plus 0.5em minus 0.4em\relax IEEE Press,
  2015, pp. 19--25.

\bibitem{squillero2016divergence}
G.~Squillero and A.~Tonda, ``Divergence of character and premature convergence:
  A survey of methodologies for promoting diversity in evolutionary
  optimization,'' \emph{Information Sciences}, vol. 329, 2016.

\bibitem{deb2000fast}
K.~Deb, S.~Agrawal, A.~Pratap, and T.~Meyarivan, ``A fast elitist non-dominated
  sorting genetic algorithm for multi-objective optimization: Nsga-ii,'' in
  \emph{International Conference on Parallel Problem Solving From
  Nature}.\hskip 1em plus 0.5em minus 0.4em\relax Springer, 2000, pp. 849--858.

\bibitem{laumanns2002combining}
M.~Laumanns, L.~Thiele, K.~Deb, and E.~Zitzler, ``Combining convergence and
  diversity in evolutionary multiobjective optimization,'' \emph{Evolutionary
  computation}, vol.~10, no.~3, pp. 263--282, 2002.

\bibitem{segurax2016using}
C.~Segura, C.~A.~C. Coello, G.~Miranda, and C.~Le{\'o}n, ``Using
  multi-objective evolutionary algorithms for single-objective constrained and
  unconstrained optimization,'' \emph{Annals of Operations Research}, vol. 240,
  no.~1, pp. 217--250, 2016.

\bibitem{vavak1996comparison}
F.~Vavak and T.~C. Fogarty, ``Comparison of steady state and generational
  genetic algorithms for use in nonstationary environments,'' in \emph{Proc. of
  IEEE Internat. Conference on Evolutionary Computation}.\hskip 1em plus 0.5em
  minus 0.4em\relax IEEE, 1996.

\bibitem{bredeche2009line}
N.~Bredeche, E.~Haasdijk, and A.~Eiben, ``On-line, on-board evolution of robot
  controllers,'' in \emph{International Conference on Artificial Evolution
  (Evolution Artificielle)}.\hskip 1em plus 0.5em minus 0.4em\relax Springer,
  2009, pp. 110--121.

\bibitem{kinneer2018managing}
C.~Kinneer, Z.~Coker, J.~Wang, D.~Garlan, and C.~Le~Goues, ``Managing
  uncertainty in self-adaptive systems with plan reuse and stochastic search,''
  in \emph{Proceedings of the 13th International Symposium on Software
  Engineering for Adaptive and Self-Managing Systems (SEAMS)}, 2018.

\bibitem{horn1994niched}
J.~Horn, N.~Nafpliotis, and D.~E. Goldberg, ``A niched pareto genetic algorithm
  for multiobjective optimization,'' in \emph{Evolutionary Computation, 1994.
  IEEE World Congress on Computational Intelligence., Proceedings of the First
  IEEE Conference on}.\hskip 1em plus 0.5em minus 0.4em\relax IEEE, 1994.

\bibitem{wolpert1997no}
D.~H. Wolpert and W.~G. Macready, ``No free lunch theorems for optimization,''
  \emph{IEEE Trans. on Evolutionary Comp.}, vol.~1, no.~1, 1997.

\bibitem{holzl2016continuous}
M.~H{\"o}lzl and T.~Gabor, ``Continuous collaboration for changing
  environments,'' in \emph{Transactions on Foundations for Mastering Change
  I}.\hskip 1em plus 0.5em minus 0.4em\relax Springer, 2016, pp. 201--224.

\bibitem{hart2018evolution}
E.~Hart, A.~S. Steyven, and B.~Paechter, ``Evolution of a functionally diverse
  swarm via a novel decentralised quality-diversity algorithm,'' \emph{arXiv
  preprint arXiv:1804.07655}, 2018.

\bibitem{berenos2010antagonistic}
C.~B{\'e}r{\'e}nos, K.~M. Wegner, and P.~Schmid-Hempel, ``Antagonistic
  coevolution with parasites maintains host genetic diversity: an experimental
  test,'' \emph{Proc. of the Royal Society of London B: Biological Sciences},
  2010.

\end{thebibliography}

\end{document}